\def\year{2020}\relax
\documentclass[letterpaper]{article} 
\usepackage{aaai20}  
\usepackage{times}  
\usepackage{helvet} 
\usepackage{courier}  
\usepackage[hyphens]{url}  
\usepackage{graphicx} 
\usepackage{graphicx}  
\usepackage[noend]{algorithm2e}
\usepackage{xspace}
\usepackage{enumitem}
\usepackage{amsmath,amsthm,amssymb}

\newenvironment{LIST} 
{\begin{itemize}[parsep=1ex,itemsep=0mm,topsep=1ex,leftmargin=1.5em]}
	{\end{itemize}}

\newcommand{\la}{\langle}
\newcommand{\ra}{\rangle}
\newcommand{\RAE}{\textsf{\small RAE}\xspace}

\newcommand{\RPLAN}{\textsf{\small RAEplan}\xspace}
\newcommand{\PLAN}{\textsf{\small UPOM}\xspace}

\newcommand{\lm}{\sv{Learn$\pi$}\xspace} 
\newcommand{\lh}{\sv{LearnH}\xspace}

\newcommand{\failed}{\textsf{\small failed}\xspace}

\newcommand{\sv}[1]{{\normalfont\textsf{\small #1}}} 

\newcommand{\argmax}{\textrm{argmax}}

\newtheorem{example}{Example}

\newlength{\tabsize}
\newcommand{\T}{\hspace*{\tabsize}}

\usepackage[dvipsnames]{xcolor}

\usepackage{newfloat}

\newenvironment{pcode}{%
	\setcounter{pline}{0}%
	\begin{tabular}[b]{@{\kern .8em}r@{~}l@{}l@{\,}}
		&&\\[-5.5ex] 
	}{%
	\end{tabular}%
}

\newcommand{\phead}[1]{
	\\ \multicolumn{3}{@{}l}{#1}}

\newcounter{pline}

\newcommand{\1}{\\&}

\newcommand{\pkey}[1]{\\#1:&}

\usepackage[normalem]{ulem}

\newcommand{\body}{\textrm{body}}

\makeatletter
\def\oldfootnotesize{\@setsize\oldfootnotesize{9pt}\viiipt\@viiipt} 
\makeatother

\usepackage{tabularx}

\newcommand{\CR}{{\small \sv{Fetch}}\xspace}
\newcommand{\SD}{{\small \sv{Nav}}\xspace}
\newcommand{\SR}{{\small \sv{S\&R}}\xspace}
\newcommand{\EE}{{\small \sv{Explore}}\xspace}

\newcommand{\current}{i}

\newcommand{\uctRollouts}{\textit{n$_{ro}$}}

\newcommand{\Rplantwo}{\sv{Select-Method}\xspace}

\title{Integrating Acting, Planning, and Learning in \\Hierarchical Operational Models}

\author{Sunandita Patra,\textsuperscript{\rm 1}
James Mason,\textsuperscript{\rm 1}
Amit Kumar,\textsuperscript{\rm 1}
Malik Ghallab,\textsuperscript{\rm 2}
Paolo Traverso,\textsuperscript{\rm 3} 
Dana Nau\textsuperscript{\rm 1}\\
\textsuperscript{\rm 1} University of Maryland, College Park, MD 20742, USA\\
\textsuperscript{\rm 2} LAAS-CNRS, 31077, Toulouse, France\\
\textsuperscript{\rm 3} Fondazione Bruno Kessler, I-38123, Povo-Trento, Italy\\
\{patras@, jmason12@terpmail., akumar14@terpmail.\}umd.edu, malik@laas.fr, traverso@fbk.eu, nau@cs.umd.edu
}

\begin{document}

\maketitle

\begin{abstract}
We present new planning and learning algorithms for  \RAE, the Refinement Acting Engine \cite{ghallab2016automated}. \RAE uses hierarchical operational models to perform tasks in dynamically changing environments. Our planning procedure, \PLAN, does a UCT-like search in the space of operational models in order a near optimal method to use for the task and context at hand. Our learning strategies acquire, from online acting experiences and/or simulated planning results, a mapping from decision contexts to method instances as well as a heuristic function to guide \PLAN. Our experimental results show that \PLAN and our learning strategies significantly improve \sv{RAE}'s performance in four test domains using two different metrics: efficiency and success ratio.
\end{abstract}

\section{Introduction}
\label{sec:intro}

The ``\textit{actor's view of automated planning and acting}'' \cite{ghallab2014actors} advocates a hierarchical organization of an actor's deliberation functions, with online planning throughout the acting process. Following this view, \cite{patra2019acting} 
proposed \RPLAN, a planner for the Refinement Acting Engine (\RAE) of \cite[Chap. 3]{ghallab2016automated}, and showed on test domains that it improves \RAE's efficiency and success ratio.
This approach, on which we rely, is appealing for its powerful representation and seamless integration of reasoning and acting.

\RAE's operational models are specified as a collection of hierarchical {\em refinement methods} giving alternative ways to perform tasks and react to events. A method has a {\em body} that can be any complex algorithm, without the restrictions of HTN methods. It may contain the usual programming constructs, as well as subtasks that need to be refined recursively, and primitive actions that query and may change the world nondeterministically. 
\RAE uses a collection of methods for closed-loop online decision making to perform tasks and react to events. When several method instances are available for a task, \RAE may respond purely reactively, relying on a domain specific heuristic. It may also call an online planner such as \RPLAN, to get a more informed decision.

\RPLAN offers advantages over similar planners (see  Sec. \ref{sec:soa}), but it is not easily scalable for demanding real-time applications, which require an anytime procedure supporting a receding-horizon planner.
We propose here a new planning algorithm for \RAE, which relies on a UCT-like Monte-Carlo tree search procedure called \PLAN (UCT Planner for Operational Models). It is a progressive deepening, receding-horizon anytime planner. Its scalability requires heuristics. However, while operational models are needed for acting and can be used for planning, they lead to quite complex  search spaces not easily amenable to the usual techniques for domain-independent heuristics. 

Fortunately, the above issue can be addressed with learning. 
A learning approach can be used to acquire a mapping from decision contexts to method instances, and this mapping can be used as the base case of the anytime strategy. Learning can also be used to acquire a heuristic function to guide the search. 
The contributions of this paper include:
\begin{LIST}
	\item A Monte-Carlo tree search procedure that extends UCT to a search space containing {\em disjunction} nodes, {\em sequence} nodes, and {\em statistical sampling} nodes. The search uses progressive deepening to provide an anytime planning algorithm that can be used with different utility criteria.
	\item Learning strategies to acquire, from online acting experiences and/or simulated planning results, both a mapping from decision contexts to refinement methods and a heuristic evaluation function to guide \PLAN.
	\item An approach to integrate acting, planning and learning for an actor in a dynamic environment.
\end{LIST}

These contributions are backed-up with a full implementation of \RAE and \PLAN and extensive experiments on four test domains, to characterize the benefits of two different learning modalities and compare \PLAN to \RPLAN.
We do not claim any contribution on the learning techniques {\em per se}, but on the integration of learning, planning, and acting. We use  an off-the-shelf  learning library with appropriate adaptation for our experiments.
The learning algorithms do not provide the operational models needed by the planner, but they do several other useful things.
First, they speed up the planner's search, thereby improving the actor's efficiency. Second, they enable both the planner and the actor to find better solutions, thereby improving the actor's success ratio.
Third, they allow the human domain author to write refinement methods without needing to specify a preference ordering in which the planner or actor should try those methods.

In the following sections we discuss the related work, then introduce informally the operational model representation and \RAE.  \PLAN procedure is detailed in Section ~\ref{sec:planning}. Section ~\ref{sec:integration} presents three ways in which supervised learning can be integrated with \RAE and \PLAN. In Section~\ref{sec:eval}, we describe our experiments and show the benefits of planning and learning with respect to purely reactive \RAE.

\section{Related work}
\label{sec:soa}

Most of the works that extend operational models with some deliberation mechanism do not perform any kind of learning.
This is true for \RPLAN \cite{patra2019acting,patra2018ape}, its predecessor SeRPE \cite{ghallab2016automated}, and for PropicePlan \cite{Despouys:1999va}, which brings  planning capabilities to PRS \cite{Ingrand:1996uj}.
It is also true for various approaches similar to PRS and \RAE, which provide  refinement capabilities and hierarchical models, e.g., 
\citeauthor{Verma:2005tl}  \citeyear{Verma:2005tl},
\citeauthor{Wang:1991ie} \citeyear{Wang:1991ie}, 
\citeauthor{Bohren:2011bh} \citeyear{Bohren:2011bh}, 
and for
\citeauthor{musliner2008evolution} \citeyear{musliner2008evolution},
\citeauthor{goldman2016hybrid} \citeyear{goldman2016hybrid}, 
which combine online planning and acting.

Works on probabilistic planning and Monte Carlo tree search, e.g., \cite{kocsis2006bandit}, as well as  works on sampling outcomes of actions, see, e.g.,  FF-replan \cite{yoon2007ff}, 
use descriptive models (that describe {\em what} actions do but not {\em how}) 
rather than operational models, and provide no integration of acting, learning, and planning.

Our approach shares some similarities with the work on planning by reinforcement learning (RL)
\cite{kaelbling1996reinforcement,sutton1998reinforcement,Geffner:2013to,leonetti2016synthesis,garmelo2016towards}, since we learn by acting in a (simulated) environment. However, most of the works on RL learn policies that map states to actions to be executed, and  learning is performed in a descriptive model. 

We learn how to select refinement methods in an operational model that allows for programming control constructs.
This main difference holds also with works on hierarchical reinforcement learning, see, e.g., \cite{yang2018peorl,parr1997reinforcement,ryan2002using}. 
Works on user-guided learning, see e.g., 
\cite{martinez2017relational,martinez2017relational2}, use model based RL to learn relational models, and the learner is integrated in a robot for planning with exogenous events.
Even if relational models are then mapped to execution platforms, the main difference with our work still holds: learning is performed in a descriptive model.
\cite{jevtic2018robot} uses RL for user-guided learning directly  in the specific case  of robot motion primitives.

The approach of \cite{Morisset:2008gy} addresses a problem similar to ours but specific to robot navigation. Several methods for performing a navigation task and its subtasks are available, each with strong and weak points depending on the context. The problem of choosing a best method for starting or pursuing a task in a given context is stated as a receding horizon planning in an MDP for which a model-explicit RL technique is proposed. Our approach is not limited to navigation tasks; it allows for richer  hierarchical refinement models and is combined with a powerful Monte-Carlo tree search technique. 

The Hierarchical Planning in the Now (HPN) of \cite{kaelbling_hierarchical_2011} is designed for integrating task and motion planning and acting in robotics. Task planning in HPN relies on a goal regression hierarchized according to the level of fluents in an operator preconditions. The regression is pursued until the preconditions of the considered action (at some hierarchical level) are met by current world state, at which point acting starts. Geometric reasoning is performed at the planning level (i) to test ground fluents through procedural attachement (for truth, entailment, contradiction), and (ii) to focus the search on a few suggested branches corresponding to geometric bindings of relevant operators using heuristics called geometric suggesters. It is also performed at the acting level to plan feasible motions for the primitives to be executed. HPN is correct but not complete; however when primitive actions are reversible, interleaved planning and acting is complete. HPN has been extended in a comprehensive system for handling geometric uncertainty  \cite{kaelbling_integrated_2013}. 

Similarly, the approach of \cite{wolfe_combined_2010-1} also addresses the integration of task and motion planning  problem. It uses an HTN approach. Motion primitives are assessed with a specific solver through sampling for cost and feasibility. An algorithm called SAHTN extends the usual HTN search with a bookkeeping mechanism to cache previously computed motions. In comparison to this work as well as to HPN, our approach does not integrate specific constructs for motion planning. However, it is more generic regarding the integration of planning and acting.

In \cite{colledanchise17behaviour,colledanchise17how}, Behavioural Trees (BT) are synthesized by planning. In \cite{colledanchise2019learning} BT are generated by genetic programming.
Building the tree refines the acting process by mapping the descriptive action model onto an operational model. We integrate acting, planning, and learning directly in an operational model with the control constructs of a programming language.  
Moreover, we learn how to select refinement methods, a natural and practical way to specify different ways of accomplishing a task.

Learning planning domain models has been investigated along several approaches.  In probabilistic planning, for example \cite{ross2011bayesian}, or \cite{katt2017learning}, learn a POMDP domain model through interactions with the environment, in order to plan by reinforcement learning or by sampling methods. 
In these cases, no integration with operational models and hierarchical refinements is provided. 

Learning HTN methods has also been investigated. HTN-MAKER \cite{hogg2008HTN} learns methods given a set of actions, a set of solutions to classical planning problems, and a collection of annotated tasks. This is extended for nondeterministic domains in \cite{hogg2009learning}.  \cite{hogg2010learning} integrates HTN with reinforcement learning, and estimates the expected values of the learned methods by performing Monte Carlo updates.   
The methods used in \RAE and \PLAN are  different because the operational models needed for acting may use rich control constructs rather than simple sequences of primitives as in HTNs. At this stage, we do not learn the methods but only how to chose the appropriate one.

\section{Acting with operational models}
\label{sec:acting}

In this section, we illustrate the operational model representation and present informally how \RAE works. The basic ingredients are tasks, actions and refinement methods. A method may have several instances depending on the values of its parameters.
Here are a few simplified methods from one of our test domains called \SR.

\begin{example}
	\label{ex:ee1}
	Consider a set $R$ of robots performing search and rescue operations in a  partially mapped area. The robots' job is to find  people needing help and bring them a package of supplies (medication, food, water, etc.). This domain is specified with state variables such as 
	$\sv{robotType}(r) \in $ \{UAV, UGV\}, with $r \in R$; 
	$\sv{hasSupply}(r) \in  \{\top, \bot\}$;
	\sv{loc}$(r) \in L$, a finite set of locations. A rigid relation $\sv{adjacent} \subseteq L^2$ gives the topology of the domain. 
	
	These robots can use actions such as \textsc{DetectPerson}$(r,$ camera) which detects if a person appears in images acquired by \textit{camera} of $r$, \textsc{TriggerAlarm}$(r,l)$, \textsc{DropSupply}$(r,l)$, \textsc{LoadSupply}$(r,l)$, \textsc{Takeoff}($r, l$), \textsc{Land}($r, l$), \textsc{MoveTo}($r,l$), \textsc{FlyTo}($r,l$). They can address tasks such as: \sv{survey}$(r$,\textit{area}), which makes a UAV $r$ survey in sequence the locations in area, \sv{navigate}$(r, l)$, \sv{rescue}$(r, l),$ \sv{getSupplies}($r$). 
	
	Here is a refinement method for the $\sv{survey}$ task:
	\begin{center}
		{\rm
			\begin{pcode}
				\phead{\sv{m1-survey}$(r, l)$}
				\pkey{task}{\sv{survey}$(r, l)$}
				\pkey{pre}{\sv{robotType}$(r) =$ \textit{UAV} and \sv{loc}$(r) = l$}
				\pkey{body} 
				
				\1 for all $l'$ in neighbouring areas of $l$:
				\1 \T \sv{moveTo}$(r, l')$
				\1 \T for \textit{cam} in \sv{cameras}($r$):
				\1 \T \T if \textsc{DetectPerson}($r$, \textit{cam}) $ = \top$ then:
				\1 \T \T \T  if \sv{hasSupply}($r$) then \sv{rescue}($r,l'$)
				\1 \T \T \T else \textsc{TriggerAlarm}$(r,l')$
			\end{pcode}
		}
	\end{center}
	
	The above method specifies that the UAV $r$ flies around and captures images of all neighbouring areas of location $l$. If it detects a person in any of the images, it proceeds to perform a rescue task if it has supplies; otherwise it triggers an alarm event. This event is processed (by some other method) by finding the closest \textit{UGV} not involved in another rescue operation and assigning to it a rescue task for $l'$. Before going to rescue a person, the chosen \textit{UGV} replenishes its supplies via the task \sv{getSupply}. Here are two of its refinement methods:
\smallskip

	{
		\rm
		\begin{pcode}
			\phead{\sv{m1-GetSupplies}$(r)$}
			\pkey{task}{\sv{GetSupplies}$(r)$}
			\pkey{pre}{\sv{robotType}$(r) =$ \textit{UGV}}
			\pkey{body} \sv{moveTo}$(r, $\sv{loc}$(BASE))$
			\1 \T \textsc{ReplenishSupplies}$(r)$
		\end{pcode}
		
		\begin{pcode}
			\phead{\sv{m2-GetSupplies}$(r)$}
			\pkey{task}{\sv{GetSupplies}$(r)$}
			\pkey{pre}{\sv{robotType}$(r) =$ \textit{UGV}}
			\pkey{body}
			$r_2 = \textrm{argmin}_{r\prime} \{\textrm{EuclideanDistance}(r, r\prime) \mid $
			\1\T\T\T\T\T\sv{hasMedicine}$(r\prime) = \textsc{True}\}$
			
			\1 if $r_2$ = None then \textsc{Fail}
			\1 else:
			\1 \T \sv{moveTo}$(r, loc(r_2))$
			\1 \T \textsc{Transfer}$(r_2, r)$
		\end{pcode}
	}
	\popQED{\qed}
\end{example}

We model an acting domain as a tuple $\Sigma=(S, \mathcal{T, M, A})$ where $S$ is the set of world states the actor may be in, $\mathcal{T}$ is the set of tasks and events the actor may have to deal with, $\mathcal{M}$ is the set of method templates for handling tasks or events in $\mathcal{T}$ (we get a method instance by assigning values to the free parameters of a method template), $\sv{Applicable}(s,\tau)$ is the set of method instances applicable to $\tau$ in  state $s$, $\mathcal{A}$ is the set of primitive actions the actor may perform. We let $\gamma(s,a)$ be the set of states that may be reached after performing action $a$ in state $s$.

\smallskip\noindent\textbf{Acting problem.}
The deliberative acting problem can be stated informally as follows: given $\Sigma$ and a task or event $\tau \in \mathcal{T}$, what is the ``best'' method $m \in \mathcal{M}$ to perform $\tau$ in a current state $s$.
Strictly speaking, the actor does not require a plan, i.e., an organized set of actions or a policy. It requires an online selection procedure which designates for each task or subtask at hand the best method instance for pursuing the activity in the current context. 

The current context for an incoming external task $\tau_0$ is represented via a  \textit{refinement stack} $\sigma$ which keeps track of how much further \RAE has progressed in refining $\tau_0$. The refinement stack is a LIFO list of tuples $\sigma = \langle (\tau, m, \current), \ldots, (\tau_0, m_0, \current_0)\rangle$, where $\tau$ is the deepest current subtask in the refinement of $\tau_0$, $m$ is the method instance used to refine $\tau$, $\current$ is the current instruction in $\body(m)$, with $\current=\sv{nil}$ if we haven't yet started executing $\body(m)$, and $m=\sv{nil}$ if no refinement method instance has been chosen for $\tau$ yet. $\sigma$ is handled with the usual stack \sv{push}, \sv{pop} and \sv{top} functions.

When \RAE addresses a task $\tau$, it must choose a method instance $m$ for $\tau$. Purely reactive \RAE make this choice with a domain specific heuristic, e.g., according to some a priori order of $\mathcal{M}$; more informed \RAE relies on a planner and/or on learned heuristics. Once a method $m$ is chosen, \RAE progresses on performing the body of $m$, starting with its first step. If the current step $m[i]$ is an action already triggered, then the execution status of this action is checked. If the action $m[i]$ is still running,  stack $\sigma$ has to wait, \RAE goes on for other pending stacks in its agenda, if any. If  action $m[i]$ fails, \RAE examines alternative methods for the current subtask. Otherwise, if the action $m[i]$ is completed successfully, \RAE proceeds with the \sv{next} step in method $m$.

\sv{next}$(\sigma, s)$ is the refinement stack resulting by performing $m[i]$ in state $s$, where $(\tau, m, i) = top(\sigma)$. It  advances within the body of the topmost method $m$ in $\sigma$ as well as with respect to $\sigma$. If $i$ is the last step in the body of  $m$, the current tuple is removed from  $\sigma$: method $m$ has successfully addressed $\tau$. In that case, if $\tau$ was a subtask of some other task, the latter will be resumed. Otherwise $\tau$ is a root task which has succeeded; its stack is removed from \sv{RAE}'s agenda. If $i$ is not the last step in $m$,  \RAE proceeds to the next step in the body of $m$. 
This step $j$ following $i$ in $m$ is defined with respect to the current state $s$ and the control instruction in step $i$ of $m$, if any. 

In summary, \RAE follows a refinement tree as in Figure~\ref{fig:tree}. At an action node it performs the action in the real world; if successful it pursues the next step of the current method, or higher up if it was its last step; if the action fails, an alternate method is tried. This goes on until a successful refinement is achieved, or until no alternate method instance remains applicable in the current state. Planning with \PLAN (described in the next section) searches through this space by doing simulated sampling at action nodes.

\section{\PLAN: a UCT-like search procedure}
\label{sec:planning}

\begin{figure}[!b]
	\begin{center}
		\includegraphics[width=1.0\columnwidth]{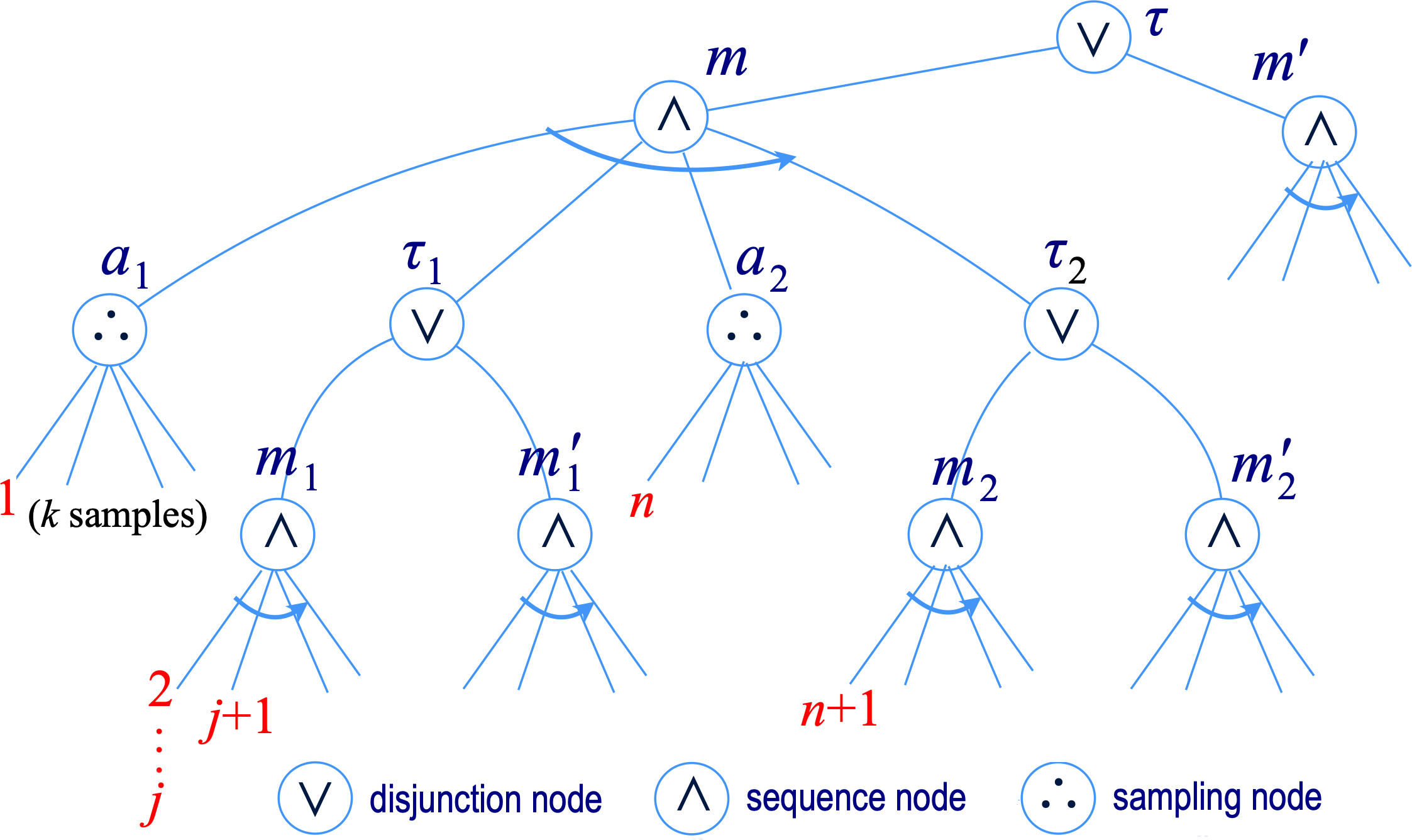}
		\caption{The space of refinement trees for a task $\tau$. A \textit{disjunction} node is a task followed by its applicable method instances. A \textit{sequence} node is a method instance $m$  followed by all the steps. A \textit{sampling} node for  an action $a$ has the possible nondeterministic outcomes of $a$ as its children. An example of a Monte Carlo rollout in this refinement tree is the sequence of nodes marked 1 (a sample of $a_1$), 2 (first step of $m_1$), $\ldots, j$ (subsequent refinements), $j+1$ (next step of $m_1$), $\ldots, n$ (a sample of $a_2$), $n+1$ (first step of $m_2$), etc. }
		\label{fig:tree}
	\end{center}
\end{figure}

\PLAN performs a recursive search to find a method instance $m$ for a task $\tau$ and a state $s$ approximately optimal for a utility function $U$. It is a UCT-like  \cite{kocsis2006bandit} Monte Carlo tree search procedure over the space of refinement trees for $\tau$ (see Figure~\ref{fig:tree}). Extending  UCT  to work on refinement trees is nontrivial since the search space contains three kinds of nodes (as shown in the figure), each of which must be handled in a different way.

\PLAN can optimize different utility functions, such as the acting efficiency or the success ratio. In this paper, we focus on optimizing the efficiency of method instances, which is the reciprocal of the total cost, as defined in \cite{patra2019acting}.

\noindent \textbf{Efficiency.} Let a method $m$ for a task $\tau$ have two subtasks, $\tau_1$ and $\tau_2$, with cost $c_1$ and $c_2$ respectively. The efficiency of $\tau_1$ is  $e_1 = 1/c_1$ and the efficiency of $\tau_2$ is $e_2 = 1/c_2$.
The cost of accomplishing both tasks is $c_1 + c_2$, so
the efficiency of $m$ is:
\begin{align}
\label{efficiency1}
1/(c_1+c_2) = 
e_1 e_2 / (e_1 + e_2).
\end{align}
If $c_1 = 0$, the efficiency for both tasks is $e_2$; likewise for $c_2 = 0$.
Thus, the incremental efficiency composition is:
\begin{align}
\label{efficiency2}
e_1 \oplus e_2 =
\mbox{}& e_2 \text{ if } e_1 = \infty,
\text{ else }\\
\nonumber
& e_1 \text{ if } e_2 = \infty, 
\text{ else }
e_1 e_2/(e_1 + e_2).
\end{align}
If $\tau_1$ (or $\tau_2$) fails, then $c_1$ is $\infty$, $e_1 = 0$. Thus 
$e_1 \oplus e_2 = 0$, meaning that $\tau$ fails with method $m$.  Note that formula \ref{efficiency2} is associative.
When using efficiency as a utility function, we denote $U$(\sv{Success}) = $\infty$ and $U$(\sv{Failure}) = 0.

\begin{algorithm}[!ht]   \DontPrintSemicolon
	\uline{$\Rplantwo(s,\tau, \sigma, d_{max}, \uctRollouts)$:} \;
	\nl \label{line:starting} $\tilde{m} \gets \text{argmax}_{m \in \sv{Applicable}(s,\tau)}h(\tau,m, s)$\; $d\gets 0$ \;
	\nl \label{slct:3}
	\Repeat{$d=d_{max}$ or \sv{searching time is over}}
	{$d \gets d+1$ \;
		\For{ $\uctRollouts$ times}
		{
			\PLAN($s$, $\sv{push}((\tau, nil, nil), \sigma$), $d$)
		}
		$\tilde{m} \gets \argmax_{m \in M} {Q_{s,\sigma}(m)}$
	}
	return $\tilde{m}$\;
	\smallskip
	\uline{$\PLAN(s, \sigma, d)$:} \;
	\lIf{$\sigma=\la \ra$}{return $U$(\sv{Success}) \text{      }} 
	
	$(\tau, m,i) \gets \sv{top}(\sigma)$\;
	\nl \lIf{$d=0$}{return $h(\tau,m, s)$}  \label{line:heuristic}

	\If { $m = nil$ or $m[i]$ is a task $\tau'$}
	{
		\lIf {$m = nil$} {$\tau' \leftarrow \tau$  \text{    \T   }\# for the first task} 
		\If { $N_{s,\sigma}(\tau')$ is not initialized yet} 
		{
			\nl \label{line:appMethods}$M' \leftarrow \sv{Applicable}(s, \tau')$\;
			\lIf {$M' = 0$} {return $U$(\sv{Failure})} 
			$N_{s,\sigma}(\tau') \leftarrow 0$ \;
			\For { $m' \in M'$}
			{
				$N_{s,\sigma}(m') \leftarrow 0$ ; 
				$Q_{s,\sigma}(m') \leftarrow 0$\;
			}
		}
		\textit{Untried} $\leftarrow \{ m' \in M' | N_{s,\sigma}(m') = 0 \}$\;
		
		\If {\textit{Untried} $ \neq \emptyset$}
		{
			$m_{chosen} \leftarrow $ random selection from \textit{Untried}
		}	\nl \label{line:ucb} 
		\lElse 
		{
			$m_{chosen} \leftarrow \argmax_{m\in M'} \phi(m,\tau')$ 
		}
		
		\nl \label{line:efftask} $\lambda \gets  \PLAN(s, \sv{push}((\tau', m,1),\sv{next}(\sigma,s)), d-1)$\;
		
		\nl \label{line:qupdate}
		$Q_{s,\sigma}(m_{chosen}) \leftarrow \frac{N_{s,\sigma}(m_{chosen}) \times Q_{s,\sigma}(m_{chosen}) + \lambda}{1 + N_{s,\sigma}({m_{chosen}})}$\;
		$N_{s,\sigma}(m_{chosen}) \leftarrow N_{s,\sigma}(m_{chosen}) + 1$\;
		return $\lambda$ \;
	}
	\If {$m[i]$ is an assignment}
	{$s' \gets$ state $s$ updated according to $m[i]$ \;
		return  $\PLAN(s', \sv{next}(\sigma,s'), d)$ } 
	
	\If {$m[i]$ is an action $a$}
	{$s'\gets \sv{Sample}(s,a)$ \;
		\lIf{$s'=\failed$}{return $U$(\sv{Failure})}	
		\nl
		\lElse{ 
			{ return $\text{$U$}(s,a,s')\oplus\PLAN(s', \sv{next}(\sigma, s'), d - 1)$ }
			\label{line:eff}}
	}
	\medskip
	\caption{\PLAN performs one rollout recursively down the refinement tree until depth $d$ for stack $\sigma$. For $C>0$, \\
		$\phi(m,\tau)=  Q_{s,\sigma}(m) +  C \sqrt{\log{N_{s,\sigma}(\tau)}/N_{s,\sigma}(m)}$.
	} 
	\label{alg:epdrUct}
\end{algorithm}

When \RAE has to perform a task $\tau$ in a state $s$ and a stack $\sigma$, it calls \Rplantwo (Algorithm~\ref{alg:epdrUct}) with two control parameters:  $n_{ro}$, the number of rollouts, and $d_{max}$, the maximum rollout length (total number of sub-tasks and actions in a rollout). \Rplantwo performs an anytime progressive deepening loop calling \PLAN $n_{ro}$ times, until the rollout length reaches $d_{max}$ or the search is interrupted. The selected method instance $\tilde{m}$ is initialized according to a heuristic $h$ (line~\ref{line:starting}). \PLAN performs recursively one Monte Carlo rollout.

When \PLAN has a subtask to be refined, it looks at the set of its applicable  method instances (line \ref{line:appMethods}).
If some method instances have not yet been tried, \PLAN chooses one randomly among \textit{Untried}, otherwise it chooses (line \ref{line:ucb}) a tradeoff between promising methods and less tried ones (Upper Confidence bound formula).  \PLAN simulates the execution of $m_{chosen}$, which may result in further refinements and actions. After the rollout is done, \PLAN updates (line \ref{line:qupdate}) the $Q$ values of $m_{chosen}$ according to its utility estimate (line \ref{line:efftask}).

When \PLAN encounters an action, it nondeterministically samples one outcome of it and, if successful, continues the rollout with the resulting state. The rollout ends when there are no more tasks to be refined or the rollout length has reached $d$. At rollout length $d$, \PLAN estimates the remaining utility using the heuristic $h$ (line \ref{line:heuristic}), discussed in Section \ref{sec:integration}. 

The planner can be interrupted anytime, which is essential for a reactive actor in a dynamic environment. It returns the method instance $\tilde{m}$ with the best $Q$ value reached so far.  For the experimental results of this paper we used fixed values of  $n_{ro}$ and $d$, without progressive deepening. The latter is not needed for the offline learning simulations.

When $d_{max}$ and $n_{ro}$ approach infinity and when there are no dynamic events, we can prove that \PLAN (like UCT) converges asymptotically to the optimal method instance for utility $U$.
Also, the $Q$ value for any method instance converges to its expected utility.\footnote{See proof at \url{https://www.cs.umd.edu/~patras/UPOM_convergence_proof.pdf}} 

\medskip
\par\noindent{\bf Comparison with \RPLAN.}
Other than UCT scoring and heuristic, \PLAN and \RPLAN \cite{patra2019acting} also differ in how the control parameters guide the search. \RPLAN does exponentially many rollouts in the search breadth, depth and samples, whereas number of \PLAN rollouts is linear in both $n_{ro}$ and $d$. \Rplantwo has more fine-grained control of the tradeoff between running time and quality of evaluation, since a change to $n_{ro}$ or $d$ changes the running time by only a linear amount.  

\section{Integrating Learning, Planning and Acting}
\label{sec:integration}

Purely reactive \RAE chooses a method instance for a task using a domain specific heuristic. \RAE can be combined with \PLAN in a receding horizon manner: whenever a task or a subtask needs to be refined, \RAE uses the approximately optimal method instance found by \PLAN. 

Finding efficient domain-specific heuristics is not easy to do by hand. 
This motivated us to try learning such heuristics automatically by running \PLAN offline in simulation over numerous cases. For this work we relied on a neural network approach, using both linear and rectified linear unit (ReLU) layers. However, we suspect that other learning approaches, e.g., SVMs, might have provided comparable results.

We have two strategies for learning neural networks to guide \RAE and \PLAN. The first one, \lm, learns a policy which maps a context defined by a task $\tau$, a state $s$, and a stack $\sigma$, to a refinement method $m$ in this context, to be chosen by \RAE when no planning can be performed. To simplify the learning process, \lm learns a mapping from contexts to methods, not to method instances, with all parameters instantiated. At acting time,   \RAE chooses randomly among all applicable instances of the learned method for the context at hand. The second learning strategy, \lh, learns a heuristic evaluation function to be used by \PLAN. 

\subsection{Learning to choose methods (\lm)}
\begin{figure}
	\centering
	\includegraphics[width=\columnwidth]{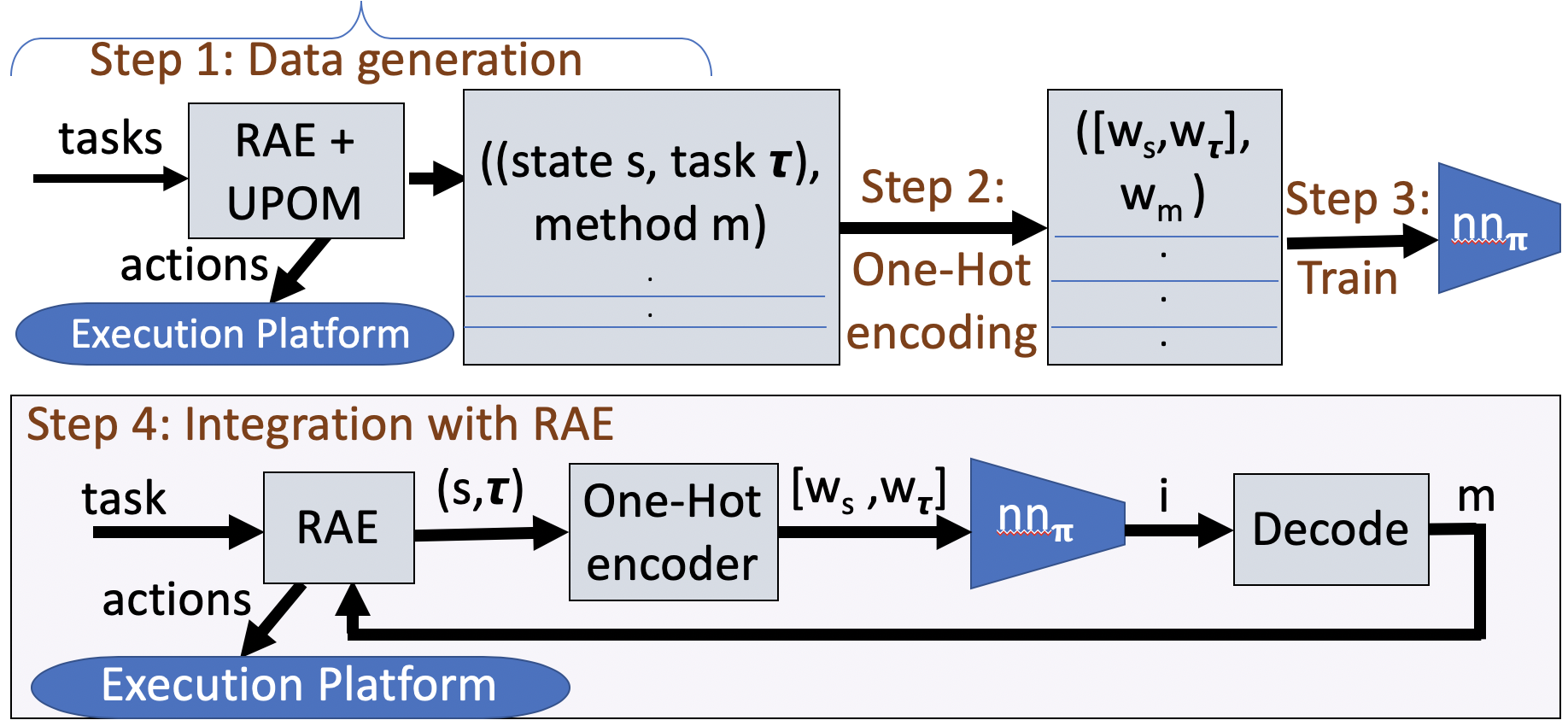}
	\caption{A schematic diagram for the \lm strategy.}
	\label{fig:lm}
\end{figure}

The \lm learning strategy consists of the following four steps, which are schematically depicted in Figure~\ref{fig:lm}. 

\smallskip\par\noindent \textbf{Step 1: Data generation.} Training is performed on a set of data records of the form $r = ((s, \tau), m)$, where $s$ is a state, $\tau$ is a task to be refined and $m$ is a method for $\tau$. Data records are obtained by making  \RAE call the planner offline with randomly generated tasks.  Each call returns a method instance $m$. We tested two approaches (the results of the tests are in Section \ref{sec:eval}):
\begin{LIST}
	\item $\lm$-1 adds $r = ((s, \tau), m)$ to the training set if \RAE succeeds with $m$ in accomplishing $\tau$ while acting in a dynamic environment.
	\item $\lm$-2 adds $r$ to the training set irrespective of whether $m$ succeeded during acting.
\end{LIST}

\smallskip\par\noindent \textbf{Step 2: Encoding.} The data records are encoded according to the usual requirements of neural net approaches. Given a record $r =((s, \tau),m)$, we encode $(s,\tau)$ into an input-feature vector and encode $m$ into an output label, with the refinement stack $\sigma$ omitted from the encoding for the sake of simplicity.\footnote{Technically, the choice of $m$ depends partly on $\sigma$. However, since $\sigma$ is a program execution stack, including it would greatly increase the input feature vector's complexity, and the neural network's size and complexity.}
Thus the encoding is
\begin{equation}
((s, \tau), m) \stackrel{\text{Encoding}}{\longmapsto} 
([w_s,w_\tau], w_m),
\label{eq:encoding}
\end{equation} 
with $w_s$, $w_\tau$ and $w_m$ being One-Hot representations of $s$, $\tau$, and $m$.
The encoding uses an $N$-dimensional One-Hot vector representation of each state variable, with $N$ being the maximum range of any state variable. Thus if every $s \in \Xi$ has $V$ state-variables, then $s$'s representation $w_s$ is $V\times N$ dimensional. 
Note that some information may be lost in this step due to discretization.

\smallskip\par\noindent \textbf{Step 3: Training.} Our multi-layer perceptron (MLP) $nn_\pi$ consists of two linear layers separated by a ReLU layer to account for non-linearity in our training data. 
To learn and classify $[w_s, w_\tau]$ by refinement methods,
we used a SGD (Stochastic Gradient Descent) optimizer and the Cross Entropy loss function.
The output of $nn_\pi$ is a vector of size $|M|$ where $M$ is the set of all refinement methods in a domain. Each dimension in the output represents the degree to which a specific method is optimal in accomplishing $\tau$.

\smallskip\par\noindent \textbf{Step 4: Integration in \RAE.}
We have \RAE use the trained network $nn_\pi$ to choose a refinement method whenever a task or sub-task needs to be refined. Instead of calling the planner, \RAE encodes $(s, \tau)$ into $[w_s, w_\tau]$ using Equation~\ref{eq:encoding}. Then, $m$ is chosen as
\begin{equation*}
m \leftarrow Decode(\argmax_i(nn_\pi([w_s, w_\tau])[i])),
\end{equation*}
where $Decode$ is a one-one mapping from an integer index to a refinement method. 

\begin{figure}
	\centering
	\includegraphics[width=\columnwidth]{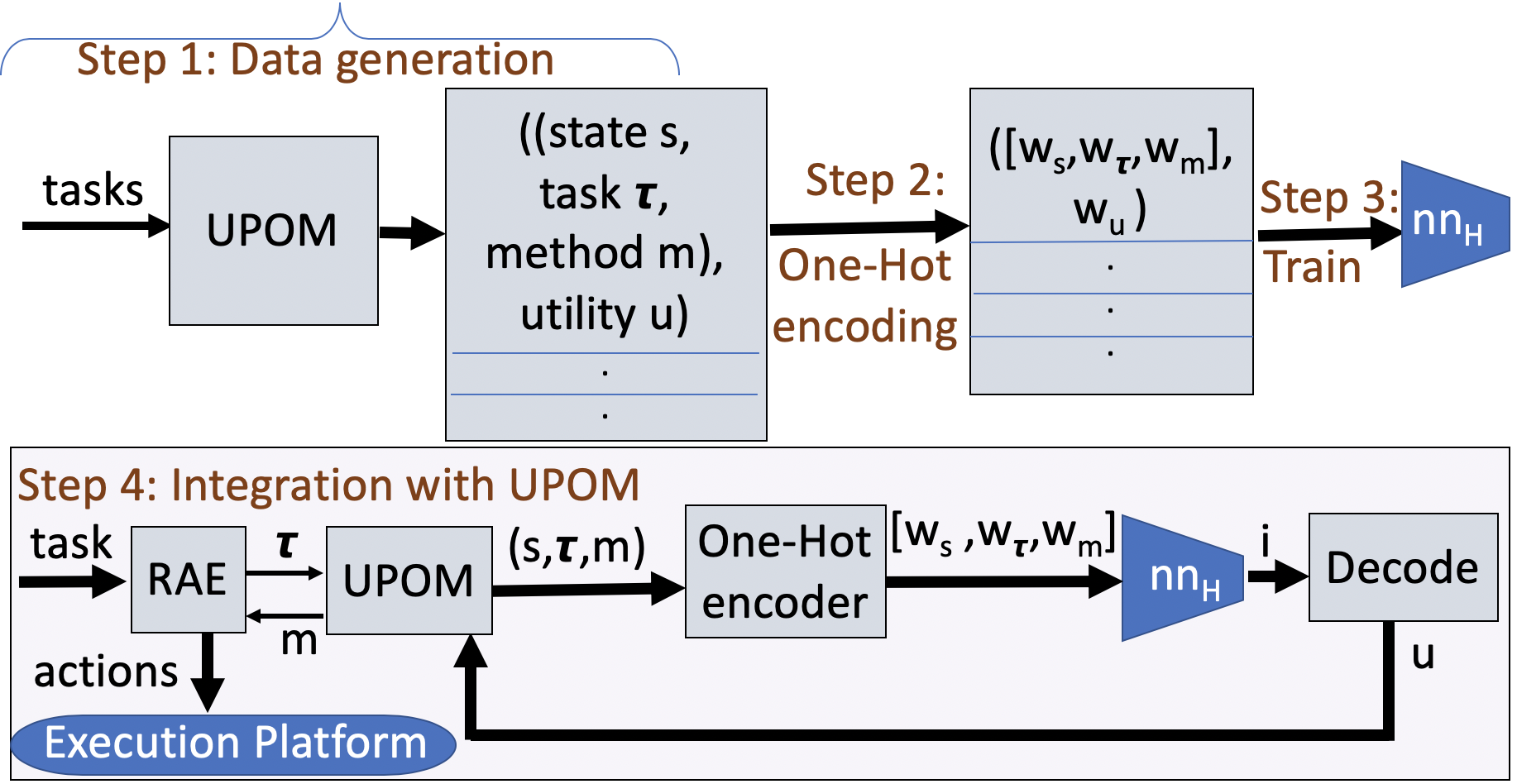}
	\caption{A schematic diagram for the \lh strategy.}
	\label{fig:lh}
\end{figure}

\begin{figure}
	\centering
	\includegraphics[width=\columnwidth]{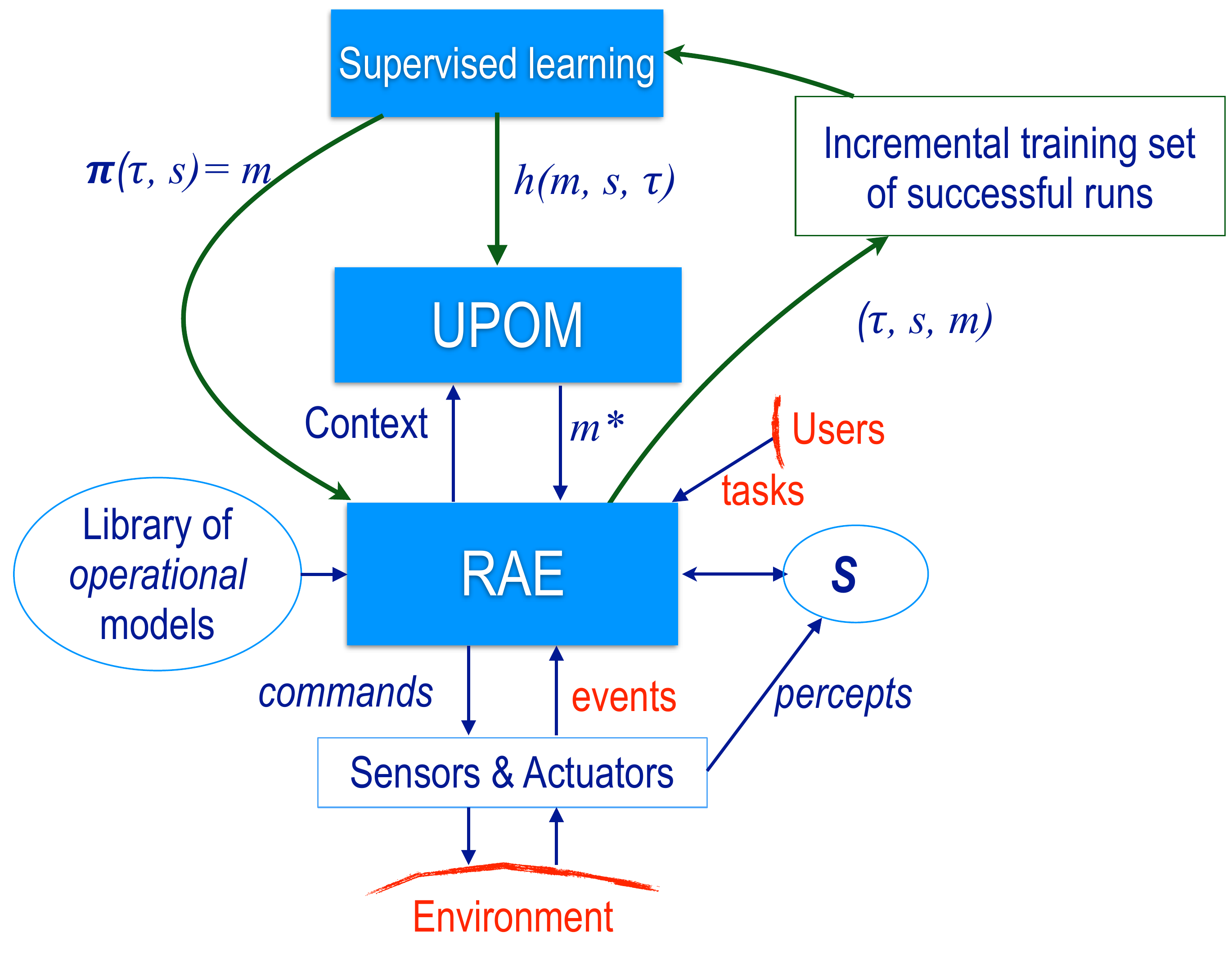}
	\caption{Integration of Acting, Planning and Learning.}
	\label{fig:archi}
\end{figure}

\subsection{Learning a heuristic function (\lh)} 

The \lh strategy tries to learn an estimate of the utility $u$ of accomplishing a task $\tau$ with a method $m$ in state $s$.
One difficulty with this is that $u$ is a real number.
In principle, an MLP could learn the $u$ values using either regression or classification. To our knowledge, there is no rule to choose between the two; the best approach depends on the data distribution. Further, regression can be converted into classification by binning the target values if the objective is discrete. In our case, we don't need an exact utility value but only need to compare utilities to choose a method. Experimentally, we observed that classification performed better than regression. 
We divided the range of utility values into $K$ intervals. By studying the range and distribution of utility values, we chose $K$ and the range of each interval such that the intervals contained approximately equal numbers of data records.
\lh learns to predict $interval(u)$, i.e., the interval in which $u$ lies.
The steps of \lh (see Figure~\ref{fig:lh}) are:

\smallskip\par\noindent \textbf{Step 1: Data generation.} We generate data records in a similar way as in the \lm strategy, with the difference that each record $r$ is of the form $((s, \tau, m), u)$ where $u$ is the estimated utility value calculated by \PLAN. 

\smallskip\par\noindent \textbf{Step 2: Encoding.}
In a record $r =((s, \tau, m), u)$, we encode
$(s, \tau, m)$ into an input-feature vector using $N$-dimensional One-Hot vector representation, omitting $\sigma$ for the same reasons as before. 
If $interval(u)$ is as described above, then the encoding is
\begin{align}
((s, \tau, m), interval(u))
&\stackrel{\text{Encoding}}{\longmapsto} ([w_s,w_\tau, w_m], w_u)
\label{eq:encodinglh}
\end{align} 
with $w_s$, $w_\tau$, $w_m$ and $w_u$ being One-Hot representations of $s$, $\tau$, $m$ and $interval(u)$.

\smallskip\par\noindent \textbf{Step 3: Training.} \lh's MLP $nn_H$ is same as \lm's, except for the output layer. $nn_H$ has a vector of size $K$ as output where $K$ is the number of intervals into which the utility values are split. Each dimension in the output of $nn_H$ represents the degree to which the estimated utility lies in that interval.

\smallskip\par\noindent \textbf{Step 4: Integration in \RAE.} 
There are two ways to use $nn_H$ with \PLAN. One is for \RAE to call the planner with a limited rollout length $d$, giving \PLAN the following heuristic function to estimate a rollout's remaining utility:
\begin{equation}
\nonumber
h(\tau, m, s) \leftarrow Decode(\argmax_{i}(nn_H([w_s, w_\tau, w_m])[i])),
\end{equation}
where $[w_s,w_\tau, w_m]$ 
is the encoding of
$(\tau, m, s)$ using Equation \ref{eq:encodinglh}, and $Decode$ is a one-one mapping from a utility interval to its mid-point.
The other way to use $nn_H$ is to estimate the heuristic function in line~\ref{line:starting} of Algorithm~\ref{alg:epdrUct}.

\subsection{Incremental online learning}

The proposed approach supports incremental online learning, although not yet implemented.
The  initialization can be performed either by running \RAE{+}\PLAN online with $d=\infty$ without a heuristic, or with an initial heuristic from offline learning on simulated data. The online acting, planning and incremental learning is performed as follows:
\begin{LIST}
	\item Augment the training set by recording successful methods and $u$ values; train the models using \lm and \lh with $Z$ records, and then switch  \RAE to use either \lm alone when no search time is available, or \PLAN with current heuristic $h$ and finite $d_{max}$ when there is some time available for planning. 
	\item Repeat the above steps every $X$ runs (or on idle periods) using the most recent $Z$ training records (for $Z$ about a few thousands) to improve the learning on both \lh and \lm.
	
\end{LIST}

\section{Experimental Evaluation}
\label{sec:eval}

\textbf{Domains.}
We have implemented and tested our framework on four simulated acting and planning domains (see Table~\ref{fig:dom_prop}).

\begin{table}[h]
	\centering
	\small
	\begin{tabular}{|@{~}c@{~~}|@{~~}c@{~~}c@{~~}c@{~}@{~~}c@{~~}c@{~~}c@{~~}c@{~~}c@{~}|}
		\hline
		Acting \&		&  			& 				&			  & Exo-	& Dead	& Sen-	& Agent	& Par-\\
		planning		&  $|\mathcal{T}|$& $|\mathcal{M}|$ & $|\mathcal{A}|$ & genous	& ends	& sing	& collab-	& allel\\
		domain	&  			& 				&			  & events		& 		& 		& oration	& tasks\\
		\hline
		\\[-2ex]
		\CR & 7 & 10 & 9 & \checkmark & \checkmark & \checkmark & -- & \checkmark\\
		\EE & 9 & 17 & 14 & \checkmark & \checkmark & -- & \checkmark & \checkmark\\
		\SD & 6 & 9 & 10 &  -- & -- & \checkmark & \checkmark & \checkmark\\
		\SR & 8 & 16 & 14 & \checkmark & \checkmark & \checkmark & \checkmark & \checkmark \\
		\hline
	\end{tabular}
	
	\caption{Properties of our domains}
	\label{fig:dom_prop}
\end{table}

In \CR, several robots are collecting objects of interest. The robots are rechargeable and may carry the charger with them.
They can't know where objects are, unless they do a sensing action at the object's location. They must search for an object before collecting it. A task reaches a dead end if a robot is far away from the charger and runs out of charge. While collecting objects, robots may have to attend to some emergency events happening in certain locations.

The \SD domain has several robots trying to move objects from one room to another in an environment with a mixture of spring doors (which close unless they're held open) and ordinary doors. A robot can't simultaneously carry an object and hold a spring door open, so it must ask for help from another robot. A free robot can be the helper. The type of each door isn't known to the robots in advance. 

The \SR domain extends the search and rescue setting of Example~\ref{ex:ee1} with UAVs surveying a partially mapped area and finding injured people in need of help. UGVs gather supplies, such as, medicines, and go to rescue the person. Exogenous events are weather conditions and debris in  paths.

In \EE, several chargeable robots with different capabilities  (UGVs and UAVs) explore a partially known terrain and gather information by surveying, screening, monitoring. They need to go back to the base regularly to deposit data or to collect a specific equipment. Appearance of animals simulate exogenous events.

\CR , \SD and \SR have sensing actions.
\CR, \SR and \EE can have dead-ends, but \SD has none.\footnote{%
	Full code is online at $\la$https://bitbucket.org/sunandita/upom/$\ra$.}

\subsection{Evaluation of planning with \PLAN}

To test whether planning with \PLAN is beneficial for \RAE, we compare its performance with purely reactive \RAE and with the planner \RPLAN
\cite{patra2019acting} in our four simulated domains.\footnote{
	We didn't compare \PLAN with any non-hierarchical planning algorithms because it would be very difficult to perform a fair comparison, as discussed in \cite{kambhampati2003are}.
}
We configured \PLAN to optimize the efficiency as its utility function, the same as \RPLAN. 

We created a test suite of 50 randomly generated problems for each domain. Each test problem consists of one to three tasks which arrive randomly chosen time points in \RAE's input stream. For each test problem, we used a maximum time limit of 5 minutes for each call to the planner. We set $n_{ro}$, the maximum number of UCT rollouts of \PLAN  to be 1000, with $d_{max} = \infty$ in each rollout.\footnote{
	The table of $N$ and $Q$ is sparse in the lower parts of the search tree but pretty dense at the top, as in the standard UCT algorithm. Each time \RAE wants to make a decision, \PLAN reruns the MCT search starting at the current node, so there's no danger of the table becoming more sparse as \RAE proceeds. $N_{s,\sigma}(m)$ is approximately in the range [50,200] at the top. In our experimental domains, the upper part of the search tree has a greater influence on the optimality, so $n_{ro}=1000$ is found to be sufficient.
}
We ran each problem 20 times, to cover sufficiently  the  non-deterministic effects of actions. We ran the tests on a 2.8 GHz Intel Ivy Bridge processor. 

\begin{figure}[h]
	\centering
	\includegraphics[width=0.9\columnwidth]{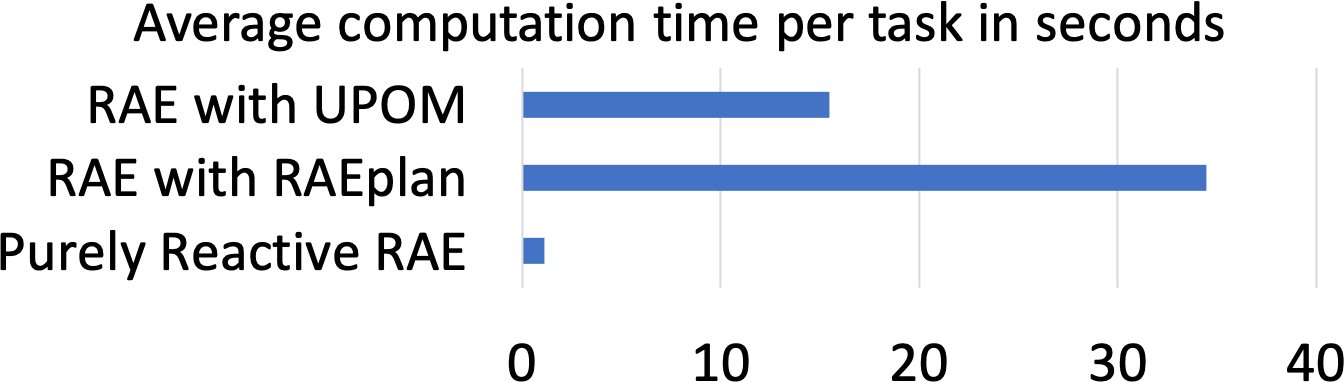}
	\caption{Computation time in seconds for a single run of a task, for \RAE with and without the planners, averaged over all four domains, an average of about $10^4$ runs. 
		If \RAE were running in the real world, the total time would be the computation time plus the time needed to perform the actions.
	}
	\label{fig:totalTime}
\end{figure}

Figure~\ref{fig:totalTime} shows the computation time for a single run of a task, averaged across all domains, an average of about $10^4$ runs (4 domains $\times $ 50 problems/domain $\times$ 1-2 tasks/problem $\times$ 20 runs/task). We observe that \RAE with \PLAN runs more than twice as fast as \RAE with \RPLAN.

\begin{figure}[t]
	\centering
	\includegraphics[width=\columnwidth]{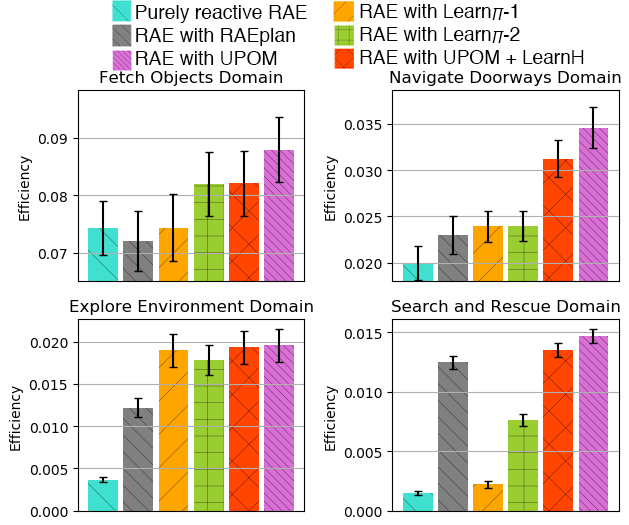}
	\caption{Efficiency (1/cost) for four domains each with six different ways of acting: purely reactive \RAE, \RAE calling \RPLAN, \RAE using the policy and heuristic learned by \lm and \lh, and \RAE using \PLAN.}
	\label{fig:nu}
\end{figure}

\smallskip\noindent \textbf{Efficiency.} The average efficiency values for all four domains are presented in Figure~\ref{fig:nu}, with the error bars showing a 95\% confidence interval.
We conclude that \RAE with \PLAN is more efficient than purely reactive \RAE and \RAE with \RPLAN
with 95\% confidence in all four domains.

\smallskip\noindent \textbf{Success ratio.} The success ratio is the proportion of incoming tasks successfully accomplished in each domain. 
Although \RPLAN and \PLAN both were configured to optimize efficiency rather than success, the success ratio is useful as a measure of robustness and is not directly proportional to efficiency. Suppose $m_1$ is always successful but has a very large cost, whereas $m_2$ sometimes fails but costs very little when it works. Then $m_1$ will have a higher probability of success, but $m_2$ will have higher expected efficiency.

\begin{figure}[t]
	\centering
	\includegraphics[width=\columnwidth]{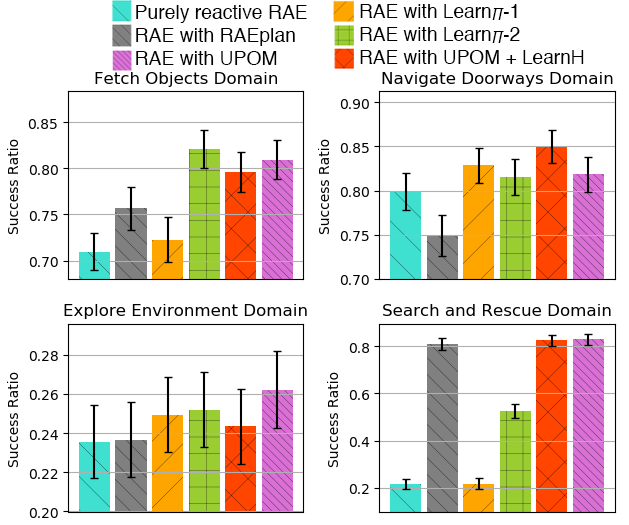}
	\caption{Success ratio (number of successful tasks / total number of incoming tasks) for four domains each with six different ways of acting: purely reactive \RAE, \RAE calling \RPLAN, \RAE using the policy and heuristic learned by \lm and \lh, and \RAE using \PLAN.}
	\label{fig:sr}
\end{figure}

Figure~\ref{fig:sr} shows \RAE's success ratio both with and without the planners.
We observe that planning with \PLAN outperforms purely reactive \RAE in \CR and \SR with 95\% confidence in terms of success ratio, whereas in \EE and \SD it does so with 85\% confidence. Also, planning with \PLAN outperforms planning with \RPLAN in \CR and \SD domains with a 95\% confidence; in \EE domain with 85\% confidence. The success ratio achieved is similar for \RPLAN and \PLAN in the \SR domain.

Asymptotically, \PLAN and \RPLAN should have near-equivalent efficiency and success ratio metrics. They differ because neither are able to traverse the entire search space due to computational constraints. Our experiments on simulated environments suggest that \PLAN is more effective than \RPLAN when called online with real-time constraints.

\begin{table*}[h]
	\begin{center}
		\resizebox{0.95\textwidth}{!}{
		\begin{tabular}{| c | c | c | c | c | c  | c | c  |  c | c | l}
			\cline{1-10}
			\text{Domain} & \multicolumn{3}{ c |}{Training Set Size}  & \multicolumn{2}{ c |}{\#(input features)}  &  \multicolumn{2}{ c |}{Training epochs}  & \multicolumn{2}{ c |}{\#(outputs)}&Note:  \\ 
			\cline{2-10}
			& \text{LM-1} & \text{LM-2} & \text{LH} &  \text{LM-1 and -2} & \text{LH} & \text{LM-1 and -2} & \text{LH} & LM-1 and -2 &  LH & LM-1 = \lm-1\\
			\cline{1-10}
			\CR  & 262 & 508 &  1084 & 97  &  104 & 430  & 250 & 10 &  100 &LM-2 = \lm-2\\ 
			\EE  & 2391 &  6883 &  10503 & 182 & 204 & 1000  & 250 & 17 & 200 & LH = \lh\\ 
			\SD  & 1686 &  5331 &  16251& 126 & 144 & 750  & 150 & 9 & 75 \\ 
			\SR  & 250 &  634 &  3542 & 330  & 401 & 225 & 250 & 16 & 10 \\ 
			\cline{1-10}
			
		\end{tabular}
}
	\end{center}
	\caption{The size of the training set, number of input features and outputs, and the number of training epochs for three different learning strategies: \lm-1, \lm-2, and \lh.}
	\label{fig:learningInfo}
\end{table*}

\subsection{Evaluation of the learning benefits}

We obtained data records for each domain by randomly generating incoming tasks and then running \RAE with \PLAN. The number of randomly generated tasks in \CR, \EE, \SD and \SR domains are 123, 189, 132 and 96 respectively.  We save the data records according to the \lm-1, \lm-2 and \lh strategies, and encode them using the One-Hot schema. We divide the training set randomly into two parts: 80\% for training and 20\% for validation to avoid overfitting on the training data. 

\begin{figure}[h]
	\centering
	\includegraphics[width=\columnwidth]{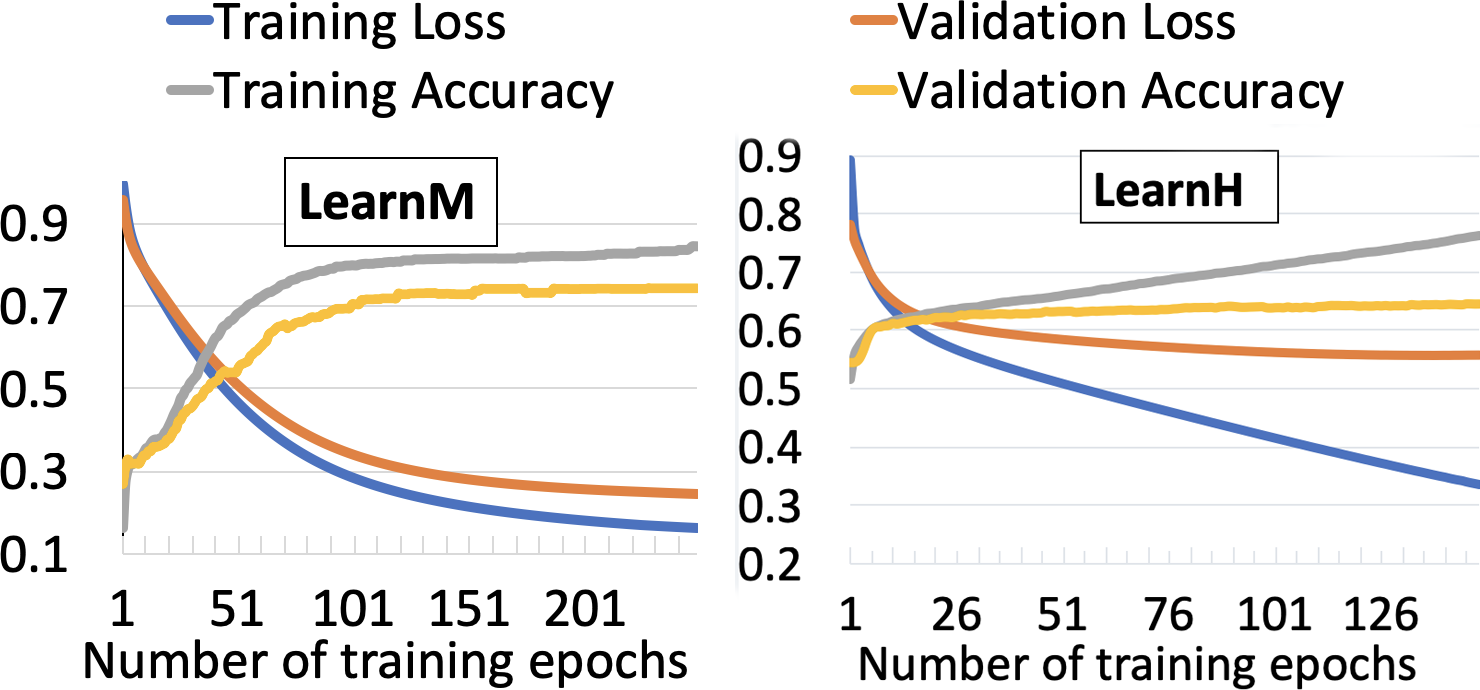}
	\caption{Training and validation results for \lm and \lh, averaged over all domains.}
	\label{fig:training}
\end{figure}

The training and validation losses decrease and the accuracies increase with increase in the number of training epochs
(see Figure~\ref{fig:training}).
The accuracy of \lm is measured by checking whether the refinement method instance returned by \PLAN matches the template predicted by the MLP $nn_\pi$, whereas the accuracy of \lh is measured by checking whether the efficiency estimated by \PLAN lies in the interval predicted by $nn_{H}$. We chose the learning rate to be in the range $[10^{-3}, 10^{-1}]$. Learning rate is a scaling factor that controls how weights are updated in each training epoch via backpropagation.
Table~\ref{fig:learningInfo} summarizes the training set size, the number of input features and outputs after data records are encoded using the One-Hot schema, number of training epochs for the three different learning strategies. In the \lh learning strategy, we define the number of output intervals $K$ from the training data such that each interval has an approximately equal number of data records. The final validation accuracies for \lm are 65\%, 91\%, 66\% and 78\% in the domains \CR, \EE, \SR and \SD respectively. The final validation accuracies for \lh are similar but slightly lower. The accuracy values may possibly improve with more training data and encoding the refinement stacks as part of the input feature vectors.  

To test the learning strategies (presented in Section~\ref{sec:integration}) we
have \RAE use $nn_{\pi}$-1, $nn_{\pi}$-2 (the models learned by \lm-1 and \lm-2) without a planner, and \RAE use \PLAN + $nn_{H}$ (the model learned by \lh), and measure the efficiency and success ratio. We use the same test suite as in our experiments with \RAE using \RPLAN and \PLAN, and do 20 runs for each test problem. When using \PLAN with $nn_H$, we set $d_{max}$ to 5 and $n_{ro}$ to 50, which has $ \sim$88\% less computation time compared to using \PLAN  with infinite $d_{max}$ and $n_{ro} = 1000$. Since the learning happens offline, there is almost no computational overhead when \RAE uses the learned models for online acting.

\smallskip\noindent \textbf{Efficiency.}  Figure~\ref{fig:nu} shows that \RAE with \PLAN+ $nn_H$ is more efficient than both purely reactive \RAE and \RAE with \RPLAN in three domains (\EE, \SR and \SD)  with 95\% confidence, and in the \CR domain with 90\% confidence. The efficiency of \RAE with
$nn_{\pi}$-1 and $nn_{\pi}$-2 lies in between \RAE with \RPLAN and \RAE with \PLAN + $nn_H$, except in the \SR domain, where they perform worse than \RAE with \RPLAN but better than purely reactive \RAE. This is possibly because the refinement stack plays a major role in the resulting efficiency  in the \SR domain. 

\smallskip\noindent \textbf{Success ratio.} In our experiments, \PLAN optimizes for the efficiency, not the success ratio. It is however interesting to see how we perform for this criteria even when it is not the chosen utility function. 
In Figure~\ref{fig:sr}, we observe that \RAE with \PLAN + $nn_H$ outperforms purely reactive \RAE and \RAE with \RPLAN in three domains (\CR, \SD and \SR) with 95\% confidence in terms of success ratio. In \EE, there is only slight improvement in success-ratio possibly because of high level of non-determinism in the domain's design. 

In most cases, we observe that \RAE does better with $nn_{\pi}$-2 than with $nn_{\pi}$-1. Recall that the training set for \lm-2 is created with all method instances returned by \PLAN regardless of whether they succeed while acting or not, whereas \lm-1 leaves out the methods that don't. This makes \lm-1's training set much smaller. In our simulated environments, the acting failures due to totally random exogenous events don't have a learnable pattern, and a  smaller training set makes \lm-1's performance worse.

\section{Conclusion}
\label{sec:conclusion}

In this paper, we have presented algorithms to guide the  acting procedure \RAE on what methods to use. One is
the \PLAN procedure, which uses a search strategy inspired by the UCT algorithm, extended to operate in a more complicated search space.
The others are learning functions:
\lm, which learns a mapping from a task in a given context to a good method,
and \lh, which provides a domain independent strategy to learn a heuristic function in a task-based hierarchical operational model framework. 

Recall that \RAE can either run purely reactively, or can get advice from an online planner. In place of the planner, we experimented using \PLAN, \RPLAN  \cite{patra2019acting}, and the models learned by \lh and \lm,
on four simulated planning-and-acting domains.
Our results show with 95\% confidence that when \RAE uses either \PLAN or the \lh model, it accomplishes tasks more efficiently than when it uses \RPLAN or runs reactively.

Furthermore, with 90\% confidence, when \RAE uses \PLAN and/or the functions \lh and \lm, its success ratio (proportion of incoming tasks accomplished successfully) is higher than when it runs reactively, even though the success ratio was not the utility function \PLAN and the learners were trying to optimize.

\smallskip\par\noindent \textbf{Future Work.} 
A significant limitation of \lm and \lh is that they give method to use, not a method {\em instance}. Thus if  they advise \RAE to use method $m$, and several different instances of $m$ are applicable in the current context, \RAE chooses among them randomly.
In future work, we may extend \lm and \lh to give advice about method instances.
Our final validation accuracy for the learning strategies is around 70\%, which shows a large scope for improvement. 

\PLAN, just like \RPLAN, uses efficiency (1/cost) as the utility function to optimize. \PLAN can easily work with other utility functions. Theoretically, \lh should also be able to estimate any utility function, but the properties of the utility function  may affect how hard it is to learn, and
we hope to test this empirically in our future work.

Currently, \lh and \lm learn offline, by calling \PLAN with randomly generated tasks. In future work, we intend to develop online learning to update the learned models while \RAE is acting.

\smallskip\par\noindent \textbf{Acknowledgement.} This work has been supported in part by NRL grant N00173191G001. The information in this paper does not necessarily reflect the position or policy of the funders, and no official endorsement should be inferred.

{\fontsize{9.9pt}{10.9pt}\selectfont
\bibliography{UPOM}
\bibliographystyle{aaai}
} 
\end{document}